\documentclass[10pt,twocolumn,letterpaper]{article}

\usepackage[pagenumbers]{wacv} 

\usepackage{graphicx}
\usepackage{amsmath}
\usepackage{amssymb}
\usepackage{booktabs}
\usepackage{multirow}
\usepackage{xcolor,colortbl}   
\usepackage{pifont}   
\usepackage[ruled,vlined]{algorithm2e}
\usepackage{rotating}
\usepackage{float}

\definecolor{Green}{RGB}{0, 180, 0}
\definecolor{Red}{RGB}{180, 0, 0}
\definecolor{Blue}{RGB}{30, 0, 180}
\definecolor{Gray}{gray}{0.9}
\definecolor{Textgray}{gray}{0.4}
\newcommand{\cmark}{{\textcolor{Green}{\ding{51}}}}%
\newcommand{\xmark}{{\textcolor{Red}{\ding{55}}}}%

\definecolor{wacvblue}{rgb}{0.21,0.49,0.74}
\usepackage[pagebackref,breaklinks,colorlinks,allcolors=wacvblue]{hyperref}

\usepackage[capitalize]{cleveref}
\crefname{section}{Sec.}{Secs.}
\Crefname{section}{Section}{Sections}
\Crefname{table}{Table}{Tables}
\crefname{table}{Tab.}{Tabs.}

\def\wacvPaperID{1964} 
\def\confName{WACV}
\def\confYear{2026}

\begin{document}

\title{Relative Energy Learning for LiDAR Out-of-Distribution Detection}

\author{
Zizhao Li \quad
Zhengkang Xiang \quad
Jiayang Ao \quad
Joseph West \quad
Kourosh Khoshelham \\
The University of Melbourne 
}
\maketitle

\begin{abstract}
   Out-of-distribution (OOD) detection is a critical requirement for reliable autonomous driving, where safety depends on recognizing road obstacles and unexpected objects beyond the training distribution. Despite extensive research on OOD detection in 2D images, direct transfer to 3D LiDAR point clouds has been proven ineffective. Current LiDAR OOD methods struggle to distinguish rare anomalies from common classes, leading to high false-positive rates and overconfident errors in safety-critical settings. We propose Relative Energy Learning (REL), a simple yet effective framework for OOD detection in LiDAR point clouds. REL leverages the energy gap between positive (in-distribution) and negative logits as a relative scoring function, mitigating calibration issues in raw energy values and improving robustness across various scenes. To address the absence of OOD samples during training, we propose a lightweight data synthesis strategy called Point Raise, which perturbs existing point clouds to generate auxiliary anomalies without altering the inlier semantics. Evaluated on SemanticKITTI and the Spotting the Unexpected (STU) benchmark, REL consistently outperforms existing methods by a large margin. Our results highlight that modeling relative energy, combined with simple synthetic outliers, provides a principled and scalable solution for reliable OOD detection in open-world autonomous driving.
   
\end{abstract}

\section{Introduction}

\begin{figure}[t]
    \centering
    \includegraphics[width=0.95\linewidth]{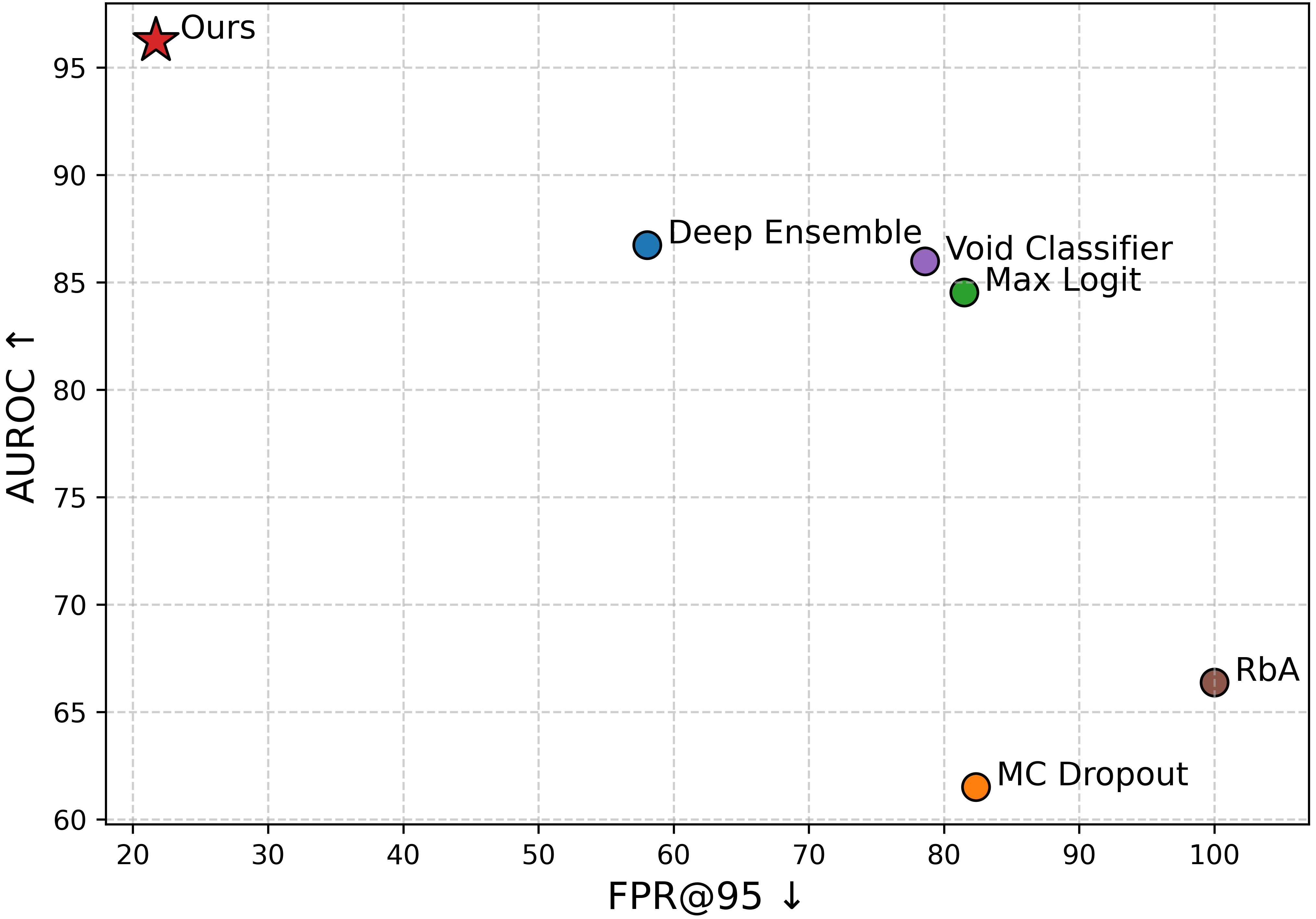}
    \caption{Point-level OOD detection performance on the STU test set. Each method is positioned by its AUROC (↑) and FPR@95 (↓). Our method achieves state-of-the-art performance, with the highest AUROC and lowest FPR@95.}
    \label{fig:ood_scatter}
    \vspace{-0.5cm}
\end{figure}


LiDAR-based perception is a cornerstone of autonomous driving, providing dense 3D information about complex road scenes. 
However, such systems must operate in open-world environments, where unexpected objects can appear at any moment. 
Out-of-distribution (OOD) objects, such as road debris, construction equipment, or other unusual obstacles, are diverse in appearance yet occur infrequently.
Their scarcity makes it infeasible to exhaustively cover them in training datasets, while their presence poses serious safety hazards: if an OOD object is confidently misclassified as an inlier, the vehicle may fail to react appropriately, potentially leading to accidents~\cite{yang2024generalizedood}. 
Therefore, robust OOD detection is indispensable for reliable LiDAR perception in autonomous driving.

Although extensively studied in 2D vision~\cite{grcic2022densehybrid,rpl2023,nayal2024likelihood,rai2023mask2anomaly,nayal2023rba,choi2023balanced,tian2022pebal}, \emph{OOD detection on LiDAR point clouds remains underexplored}. Unlike images, LiDAR data is sparse, irregular, and distance-dependent, with anomalies often represented by only a handful of points~\cite{nekrasov2025stu}. Existing approaches, such as maximum softmax probability~\cite{hendrycks2018baseline,hendrycks2022streethazards}, uncertainty estimation~\cite{srivastava2014mcdropout}, or void classification~\cite{blum2021fishyscapes}, consistently underperform in this regime, either producing \emph{overconfident predictions on anomalies} or \emph{yielding high false positives on inliers}.  
Unsupervised OOD detection approaches~\cite{hendrycks2018baseline,hendrycks2022streethazards,nayal2023rba,md} are attractive due to their simplicity, but they often suffer from severe overconfidence on anomalies, failing to provide reliable separation between inliers and OOD objects. On the other hand, methods that leverage auxiliary OOD data, such as Outlier Exposure (OE)~\cite{hendrycks2019oe} and energy-based OOD detection~\cite{tian2022pebal,energy,rpl2023}, improve robustness but come with strong assumptions on the availability and representativeness of proxy outliers. 

Following prior experience in OOD detection, we believe that robust LiDAR OOD detection can be achieved through a two-step strategy. First, one needs a realistic OOD synthesis that respects LiDAR geometry, ensuring that auxiliary anomalies resemble physically plausible structures rather than random noise. Second, an OOD detection method must be able to effectively learn from auxiliary OOD data without harming in-distribution performance.

To address the lack of OOD examples during training, we introduce a lightweight synthesis strategy, termed \emph{Point Raise}, which perturbs point clouds to generate auxiliary pseudo-OOD samples while preserving inlier semantics.
This exposes the model to diverse negative samples and stabilizes the learning of relative energy margins.
Unlike previous methods~\cite{lion2025,mood3D}, it does not rely on external datasets as OOD sources or require post-processing for point density adjustment.

To learn robust OOD decision boundaries from limited auxiliary data, we further propose \emph{Relative Energy Learning (REL)}, a framework that optimizes the \emph{energy margin} between negative and positive class logits, facilitating more stable end-to-end learning of OOD boundaries.
Compared to traditional hinge energy methods~\cite{energy}, REL eliminates the need for manual margin tuning and complex hyperparameter selection while maintaining strong discriminative capability.

We evaluate REL on SemanticKITTI~\cite{behley2019semantickitti} and the new Spotting the Unexpected (STU) benchmark~\cite{nekrasov2025stu}. 
As shown in~\cref{fig:ood_scatter}, REL achieves state-of-the-art OOD detection, attaining the highest AUROC and the lowest FPR@95 across both datasets, over both post-hoc and learning-based baselines. 
Compared to the strongest baseline, REL reduces FPR@95 by \textbf{36\%}, which enables better decision making in safety-critical LiDAR perception.

Our key contributions are:

\begin{itemize}
    \item We propose \textbf{Relative Energy Learning (REL)}, which formulates OOD detection via the relative energy margin.  
    \item We introduce \textbf{Point Raise}, a simple auxiliary data generation strategy for synthesizing pseudo OOD point clouds.  
    \item We achieve state-of-the-art results on SemanticKITTI and STU, with notably lower false-positive rates than existing methods.
\end{itemize}

\section{Related Work}

\paragraph{OOD Detection.}
OOD detection aims to identify test samples that lie outside the training distribution, thereby allowing models to reject unknown inputs rather than making overconfident and potentially erroneous predictions~\cite{yang2024generalizedood}. In safety-critical applications such as autonomous driving, OOD detection is essential, as unexpected obstacles, road debris, or novel objects may appear without warning and must be reliably detected to ensure safe navigation.

Early work on OOD detection primarily focused on image classification tasks~\cite{hendrycks2018baseline,hendrycks2019oe,odin,energy,md}. More recently, research has expanded to dense prediction tasks, where pixel-wise OOD detection are often referred to as anomaly segmentation, has gained much attention. This shift has been facilitated by benchmarks such as Fishyscapes~\cite{blum2021fishyscapes} and SegmentMeIfYouCan~\cite{chan2021segmentmeifyoucan}. Early anomaly segmentation methods~\cite{hendrycks2018baseline,chan2021entropy,lakshminarayanan2017deepensemble,tian2022pebal,grcic2022densehybrid,liang2022gmmseg,rpl2023} are commonly built upon pixel-wise semantic segmentation networks such as DeepLab~\cite{chen2018deeplabv3}.

Recent approaches increasingly adopt MaskFormer-style architectures~\cite{cheng2021maskformer,cheng2021mask2former}, with multiple extensions~\cite{delic2024uno,nayal2023rba,rai2023mask2anomaly} adapting Mask2Former for anomaly segmentation. These architectures offer improved OOD scoring by incorporating instance-level reasoning beyond purely pixel-wise predictions. Notably, many of these methods~\cite{tian2022pebal,chan2021entropy,rpl2023,delic2024uno,nayal2024likelihood,nayal2023rba,rai2023mask2anomaly} leverage auxiliary OOD data during training, which has been shown to substantially enhance performance. A common strategy involves pasting OOD objects from COCO~\cite{lin2014coco} into in-distribution scenes in the Cityscapes dataset~\cite{cordts2016cityscapes}.

\paragraph{LiDAR OOD Detection.}
In contrast, OOD detection for 3D LiDAR data remains relatively underexplored. Early efforts~\cite{mood3D,kosel2024mmod3d} apply OOD detection methods in a post-hoc fashion to pretrained 3D object detectors, evaluating them on public driving datasets such as KITTI~\cite{geiger2013KITTI} and nuScenes~\cite{caesar2020nuscenes}, or skip the object localization step~\cite{li2024contrastive}.

For LiDAR semantic segmentation, Cen \etal~\cite{cen2021real} introduced the REdundAncy cLassifier (REAL), which adds auxiliary logits to learn pseudo-OOD objects generated by scaling point clouds. Li~\etal~\cite{apf2023} adopted adversarial prototypes for feature-level OOD learning. Xu \etal \cite{lion2025} proposed LiON, which synthesizes diverse outliers from ShapeNet~\cite{shapenet2015} and applies a point-wise abstaining penalty, effectively seperate inliers and outliers within a selective classification framework.

These methods typically reuse existing autonomous driving datasets~\cite{caesar2020nuscenes,behley2019semantickitti}, treating minority or excluded classes as pseudo-OOD samples. Such a setup does not capture real road hazards and limits the diversity of OOD instances.
More recently, Nekrasov~\etal~\cite{nekrasov2025stu} introduced the STU benchmark, extending 2D anomaly segmentation methods to 3D LiDAR data and establishing Mask4Former~\cite{yilmaz2024mask4former} as a strong baseline. The dataset simulates a diverse range of realistic road hazards and provides both point-level and object-level annotations.



\section{Method}

\subsection{Problem Definition}
OOD detection can be formulated as a binary classification problem. 
Given an input $x \in \mathcal{X}$, the detector must decide whether $x$ is drawn from the in-distribution $\mathcal{D}_{\text{in}}$ or from some unknown distribution $\mathcal{D}_{\text{out}}$. 
This is typically achieved by defining a scoring function $S(x)$ and applying a threshold $\tau$:  
\begin{equation}
    D(x) = 
    \begin{cases}
        \text{ID}, & S(x) \leq \tau, \\
        \text{OOD}, & S(x) > \tau,
    \end{cases}
\end{equation}
where in-distribution (ID) samples are expected to yield lower scores than OOD ones. For point cloud segmentation, this binary decision is made per point, with the OOD detector assigning each 3D point a score that indicates how likely it is to be ID or OOD.

\begin{figure*}[t]
    \centering
    \includegraphics[width=\linewidth]{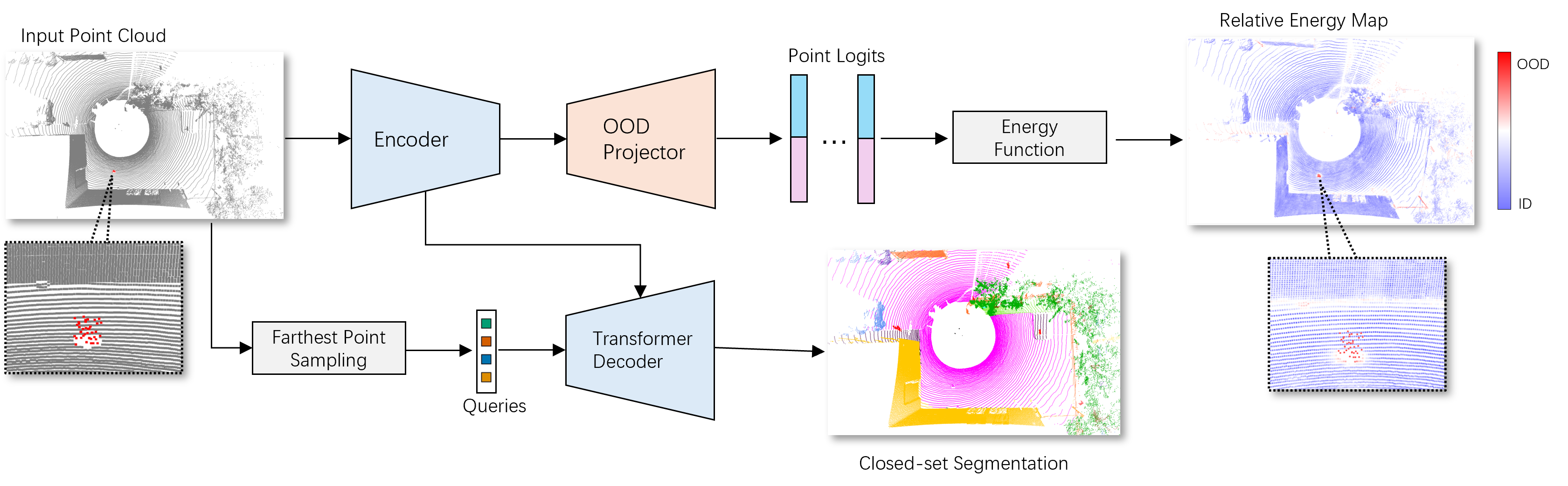}
    \caption{Overview of the Relative Energy Learning (REL) framework. We adopt the Mask4Former~\cite{yilmaz2024mask4former} backbone with a sparse UNet encoder~\cite{choy2019minkowski} and a transformer decoder operating on Farthest Point Sampled (FPS) queries~\cite{qi2017pointnetdeephierarchicalfeature,schult2023mask3d}. The transformer decoder produces closed-set segmentation masks, while in parallel, a lightweight OOD branch generates dense point-wise energy scores. The OOD branch is trained with the REL objective, and its outputs are converted into uncertainty maps for OOD detection.}
    \label{fig:framework}
\end{figure*}

\subsection{Overall Framework}
\paragraph{Overview.}
Our model is built on the Mask4Former-3D architecture~\cite{yilmaz2024mask4former}, which combines a transformer decoder with a multi-scale encoder for 3D panoptic segmentation.  
As shown in~\cref{fig:framework}, we extend this backbone with an additional \emph{OOD projector} branch, trained under the proposed Relative Energy Learning (REL) objective.  
This design enables joint learning of inlier semantic segmentation and point-wise OOD detection within a single unified model.

\paragraph{Mask-based Segmentation.}
This branch (the encoder and the transformer decoder) focuses on closed-set prediction. 
The encoder extracts multi-scale features from the LiDAR point cloud, which are decoded into mask embeddings by the transformer decoder.
Following Mask4Former~\cite{yilmaz2024mask4former}, each query predicts both a mask and a class label, forming the primary segmentation output.
The training procedure is identical to the closed-set setting in Mask4Former: predicted masks are matched to ground-truth masks with the Hungarian algorithm~\cite{Hungarian}, using a cost that combines mask overlap and classification accuracy. This branch ensures strong in-distribution segmentation performance.

\paragraph{OOD Detection Branch with Relative Energy Learning.}
Although mask-based segmentation models achieve strong performance in panoptic segmentation, we find that directly adapting them for OOD detection is challenging. The prediction of masks and class labels is decoupled, and the transformer decoder contains a large number of parameters, making it difficult to train effectively.

Therefore, we introduce a lightweight auxiliary branch, termed the OOD projector, composed of three linear layers with ReLU activations, attached directly after the point encoder. It processes per-point encoder features and outputs a $2K$-dimensional logit vector, where the first $K$ dimensions correspond to ID class logits (positive) and the remaining $K$ represent OOD surrogate logits (negative).

We compute the relative energy margin by applying the energy function to each group and taking their difference, quantifying the evidence for OOD versus ID classification. This OOD branch enables dense point-wise OOD scoring without modifying the main segmentation head, thereby preserving closed-set performance.

\paragraph{Inference.}
Finally, at test time, the segmentation predictions are obtained from the transformer decoder, while the OOD projector computes relative energy for each point.  
Points with large relative energy are flagged as OOD, while points with negative relative energy are retained as in-distribution.  
The joint design ensures that the framework outputs high-quality semantic segmentation and reliable point-wise OOD scores simultaneously.

Next, we present the details of REL, which forms the core of our OOD detection framework.


\subsection{Relative Energy Learning}
\paragraph{Preliminaries.}
Energy-based models (EBMs) associate each input-label pair $(x,y)$ with a scalar energy $E(x,y)$, where lower energy indicates higher compatibility~\cite{lecunebm}. 

For a discriminative classifier $f(x) \in \mathbb{R}^K$ producing $K$ logits, the free energy can be expressed directly from the energy function of logits~\cite{energy}:  
\begin{equation}
    E(x; f) = -T \cdot \log \sum_{i=1}^Ke^{f_i(x)/T}.
    \label{eq:energy}
\end{equation}
In this way, input data are mapped to a single scalar, which represents uncertainty.
This value is typically small for in-distribution samples and larger for OOD samples.  
Thus, an OOD detector can be constructed by thresholding the energy instead of using the maximum softmax probability.  


In addition, energy-based approaches not only use the free energy score at inference time, but also fine-tune the network to explicitly enlarge the gap between ID and OOD samples~\cite{energy,du2021vos,tian2022pebal,rpl2023,choi2023balanced,dual_energy}.
This is typically achieved via a hinge energy loss, where in-distribution instances are penalized if they exceed a margin $m_{\text{in}}$, and OOD objects are penalized if they fall below a margin $m_{\text{out}}$:  
\begin{align}
    \mathcal{L}_{\text{energy}} 
    &= \mathbb{E}_{x \sim \mathcal{D}_{\text{in}}} \left[ \max(0, E(x) - m_{\text{in}})^2 \right] \nonumber \\
    &+ \mathbb{E}_{x \sim \mathcal{D}_{\text{out}}} \left[ \max(0, m_{\text{out}} - E(x))^2 \right].
\end{align}
Although effective, this design introduces multiple hyperparameters ($m_{\text{in}}, m_{\text{out}}$, temperature scaling), requiring careful calibration on various tasks.

\paragraph{Motivation.}
Energy-based OOD detection methods~\cite{energy,du2021vos,tian2022pebal,rpl2023,dual_energy} have emerged as a simple and effective paradigm. However, their performance often depends on multiple hyperparameters such as fixed thresholds, temperature scaling, and manually defined energy margins, which reduces robustness and makes them less convenient to use. Moreover, many methods typically ignore the severe class imbalance inherent in LiDAR segmentation: a single scan could contain millions of ID points but only a handful of OOD points, making threshold calibration unreliable and biasing learning toward the dominant ID distribution. We propose \emph{Relative Energy Learning (REL)} to address these challenges. 

REL introduces a contrastive energy margin between ID and negative OOD logits, yielding a shift-invariant energy gap between ID and OOD samples, and employs an imbalance-aware loss with a single weighting parameter. Introducing negative logits helps to better partition the logit space and explicitly model OOD samples.
 This formulation eliminates the need for delicate hyperparameter tuning while explicitly leveraging scarce auxiliary OOD samples to sharpen decision boundaries under class imbalance.

\paragraph{Energy Formulation.}
We consider semantic segmentation as a point-wise classification problem. 
Given an input point $x$, the network produces logits $f(x)\in\mathbb{R}^{2K}$, 
where the first $K$ channels correspond to ID classes ($y^+$) 
and the remaining $K$ channels serve as negative OOD counterparts ($y^-$). 
We define the relative energy margin as

\begin{equation}
\Delta E(x;f) = \log \frac{\sum_{i \in y^-} \exp(f_i(x))}
           {\sum_{i \in y^+} \exp(f_i(x))}.
\label{eq:rel-margin}
\end{equation}
Thus, $\Delta E(x;f)$ is computed as the log-ratio between the aggregated OOD evidence and the aggregated ID evidence.
This generalizes the negative free-energy score in energy-based OOD detection~\cite{energy} by adding negative logits, making the decision boundary explicitly trainable.

To better understand $\Delta E$, let us first introduce the softmax normalizing constant.
Let
$Z(x) = \sum_{j=1}^{2K} \exp(f_j(x))$
be the normalizing constant of the softmax distribution. 
We then define the grouped probabilities of ID and OOD as
\begin{equation}
p_{\text{pos}}(x) = \sum_{i \in y^+} \frac{\exp(f_i(x))}{Z(x)}, 
p_{\text{neg}}(x) = \sum_{i \in y^-} \frac{\exp(f_i(x))}{Z(x)}.
\end{equation}
Thus $p_{\text{pos}}(x)$ is the total softmax probability mass assigned to all ID classes, 
and $p_{\text{neg}}(x)$ is the total mass assigned to the auxiliary OOD channels. 

In this way, the relative energy can be written as
\begin{equation}
\Delta E(x;f) = \log \frac{p_{\text{neg}}(x)}{p_{\text{pos}}(x)},
\end{equation}
which is exactly the \emph{log-odds} of ``OOD vs.\ ID'' under the grouped softmax. 
For ID points, $p_{\text{pos}}(x)\!\gg\!p_{\text{neg}}(x)$, leading to $\Delta E(x;f)\!\ll 0$. 
For OOD points, $p_{\text{neg}}(x)$ becomes comparable to or exceeds $p_{\text{pos}}(x)$, 
causing $\Delta E(x;f)$ to approach or cross zero. 
This relative energy serves as a smooth decision statistic and forms the basis of our learning objective.

\paragraph{Learning Objective.}
To enable end-to-end training with minimal complexity, we formulate OOD detection as a binary classification problem~\cite{du2021vos}. 
In addition, we minimize a logistic loss with imbalance weighting. Specifically, We define the loss function as
\begin{align}
\mathcal{L}_{\text{REL}}
&= \mathbb{E}_{x \sim \mathcal{D}_{\text{in}}}
\Big[-\log\!\Big(\frac{\exp^{-\Delta E(x;f)}}{1+\exp^{-\Delta E(x;f)}}\Big)\Big] \nonumber\\
&\quad + \omega \,\mathbb{E}_{x \sim \mathcal{D}_{\text{aux}}}
\Big[-\log\!\Big(\frac{1}{1+\exp^{-\Delta E(x;f)}}\Big)\Big],
\label{eq:rel-logistic}
\end{align}
where $\mathcal{D}_{\text{in}}$ denotes ID points, 
$\mathcal{D}_{\text{aux}}$ are auxiliary OOD points generated by Point Raise, 
and $\omega$ compensates for the scarcity of OOD data. 
This loss enlarges the ID vs.\ OOD margin by concentrating gradients on boundary cases ($\Delta E\!\approx\!0$), which yields sharper and more reliable separation under severe LiDAR class imbalance. It is also simple because it introduces only a single imbalance weight, thereby minimizing hyperparameter tuning.

\subsection{Point Raise: A Lightweight Auxiliary OOD Synthesis Algorithm}

A common strategy for OOD detection in 2D vision is \emph{Outlier Exposure (OE)}, 
where OOD samples are synthesized by cropping and pasting objects from external datasets~\cite{hendrycks2019oe,rpl2023,grcic2022densehybrid,tian2022pebal,chan2021entropy}. 
While effective for dense images, such approaches are poorly suited to LiDAR. 
Point clouds are inherently sparse, irregular, and range-dependent: directly transplanting objects introduces unrealistic artifacts, broken boundaries, and mismatched densities. 
Previous studies have attempted to extend this concept to 3D data~\cite{cen2021real,lion2025}, but such approaches must account for the point density variations and occlusion patterns inherent to LiDAR sensing.
These challenges motivate us to develop a simpler and more effective strategy for OOD sample synthesis.

We propose Point Raise, a simple augmentation that generates realistic pseudo OOD clusters directly from training scenes. 
Road surfaces provide abundant and uniformly sampled points, which can be perturbed to simulate anomalies. 
By pulling a local cluster of road points inward and then raising them vertically, Point Raise produces compact, protruding structures that preserve local density but disrupt global consistency. 
These clusters effectively simulate OOD objects of various shapes without relying on external datasets.  

As shown in~\cref{pointraise}, the proposed Point Raise algorithm first samples a random road point, gathers its neighbors within a random radius using a KD-tree, 
contracts them radially toward the sensor with an adaptive decay scaled by the pull factor $\gamma$, 
and finally perturbs their heights by random offsets in $[h_{\min},h_{\max}]$. 
The modified points are relabeled as a reserved OOD class.  

This pulling process leverages LiDAR geometry: the sensor sits at the origin, and road points form flat radial layers. 
Simply shifting road points upward would yield unrealistic slabs, whereas radial contraction transforms a patch into a compact, object-like cluster. 
Formally, let $d$ denote the distance of each cluster point from the sensor, with $d_{\min}$ and $d_{\max}$ the nearest and farthest points. 
We define $d_{\text{shift}} = d - d_{\min}$ and compute an adaptive decay
\begin{equation}
a = -\log(d_{\min}/d_{\max}) \,/\, \big(\gamma (d_{\max} - d_{\min})\big).
\end{equation}
Each point is then scaled by $s = \exp(-a \cdot d_{\text{shift}})$, 
so that the near points remain almost unchanged while the far points are pulled inward, compacting the cluster into an object-like anomaly.

\begin{algorithm}[t]
\caption{Point Raise Algorithm}
\SetCommentSty{scriptsize}
\KwIn{Point cloud $P$, labels $L$, radius range $[r_{\min}, r_{\max}]$, height range $[h_{\min}, h_{\max}]$, pull factor $\gamma$}
\KwOut{Modified $P, L$}

$c \gets \text{random road point}$\;
$r \sim \mathcal{U}(r_{\min}, r_{\max})$\tcp*{uniform sampling a radius}
$cluster \gets \text{KDTree}(P).\text{query\_ball}(c, r)$\;

$d \gets \|P[cluster]\|_2$\;
$a \gets \frac{-\log(d_{\min}/d_{\max})}{\gamma \,(d_{\max} - d_{\min})}$\;
$d_{shift} \gets d - d_{\min}$\; $s \gets \exp(-a \cdot d_{shift})$\; $P[cluster] \gets P[cluster,xy] \odot s$ \tcp*{element-wise scaling}
$h \sim \mathcal{U}(h_{\min}, h_{\max})^{|\text{cluster}|}$\tcp*{uniform sampling heights}
$P[cluster,z] \gets P[cluster,z] + h$\; 
$L[cluster] \gets \texttt{RAISED\_CLASS}$\;

\Return{$P,L$}
\label{pointraise}
\end{algorithm}

\begin{figure}[htbp]
    \centering
    \includegraphics[width=\linewidth]{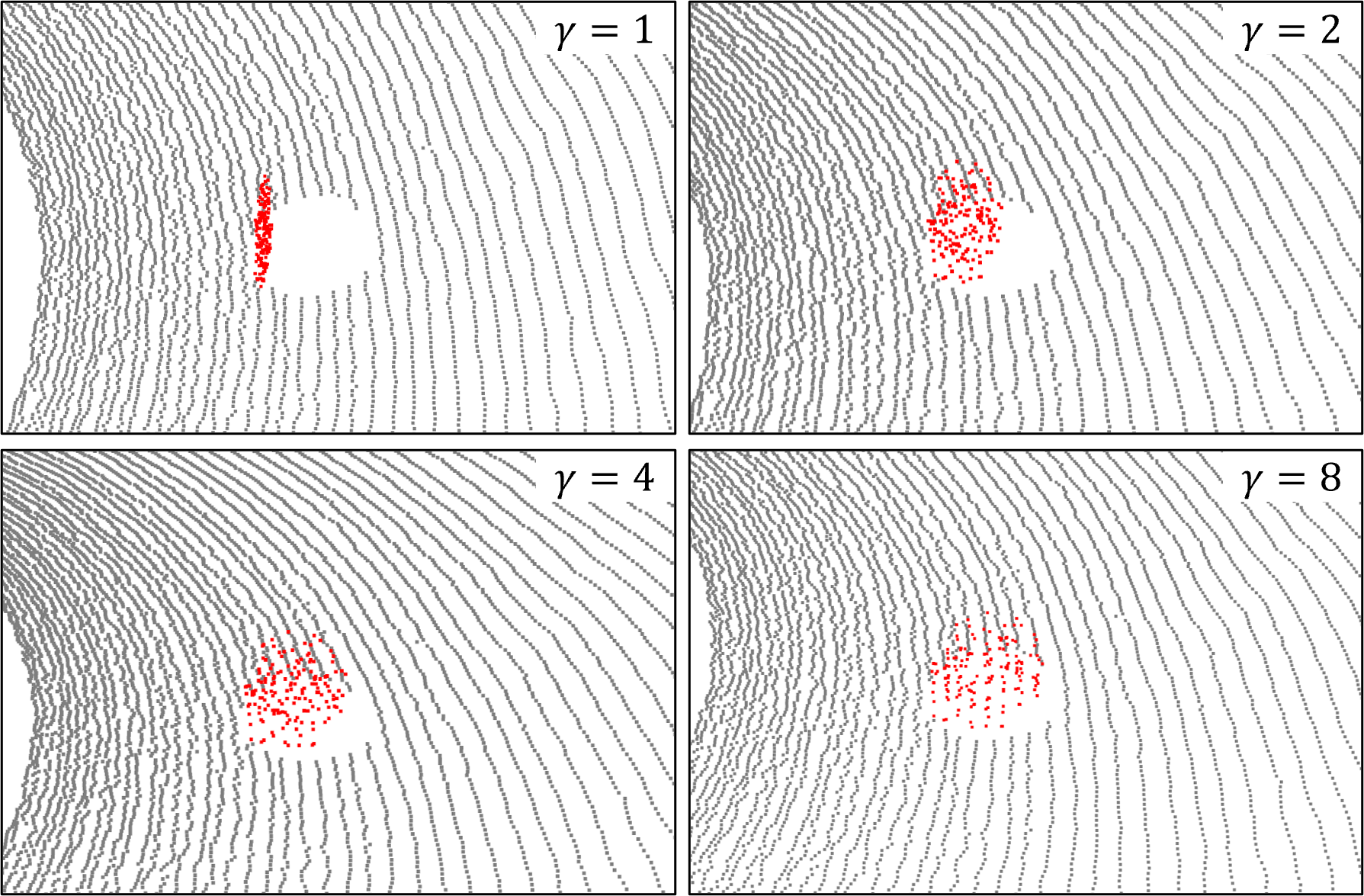}
    \caption{Visualization of Raise Cluster with Different $\gamma$.}
    \label{fig:gamma}
\end{figure}

As shown in \cref{fig:gamma}, with small values ($\gamma=1$), the cluster is strongly contracted, producing a tight, blob-like anomaly. As $\gamma$ increases ($\gamma=2,4,8$), the contraction weakens and the clusters become larger and flatter. This parameter thus controls the compactness of the synthesized anomalies, ranging from dense protrusions at low $\gamma$ to more subtle, spread-out structures at high $\gamma$.

\begin{table*}[ht]
    \centering
    \caption{Anomaly Segmentation Performance on the Validation Set of STU benchmark. * denotes the SOTA method reproduced by us. Other results are taken directly from~\cite{nekrasov2025stu}. All methods use the Mask4Former~\cite{yilmaz2024mask4former} architecture.}
    \scriptsize
    \resizebox{\textwidth}{!}{
    \begin{tabular}{lcccccccccc}
        \toprule
        \multirow{2}{*}{Method} & \multirow{2}{*}{\shortstack{Auxiliary \\ OOD Data}} & \multicolumn{3}{c}{Point-Level OOD} && \multicolumn{5}{c}{Object-Level OOD} \\
        \cline{3-5} \cline{7-11}
         & & AUROC~$\uparrow$ & FPR@95~$\downarrow$ & AP~$\uparrow$ && RecallQ~$\uparrow$ & SQ~$\uparrow$ & RQ~$\uparrow$ & UQ~$\uparrow$ & PQ~$\uparrow$ \\
        \midrule
        Deep Ensemble~\cite{lakshminarayanan2017deepensemble} & \xmark & 90.93 & 37.34 & \cellcolor{gray!20}6.94 && 17.70 & 79.96 & 9.10 & 14.15 & \cellcolor{gray!20}7.27 \\
        MC Dropout~\cite{srivastava2014mcdropout}    & \xmark       & 65.76 & 79.82 & \cellcolor{gray!20}0.17 && 3.54 & 74.36 & 3.48 & 2.63 & \cellcolor{gray!20}2.59 \\
        Max Logit~\cite{hendrycks2018baseline}      & \xmark     & 87.27 & 68.76 & \cellcolor{gray!20}2.02 && 26.64 & 79.26 & 2.06 & 21.12 & \cellcolor{gray!20}1.63 \\
        Void Classifier~\cite{blum2021fishyscapes} & \cmark & 89.77 & 79.50 & \cellcolor{gray!20}2.62 && 17.35 & \textbf{81.27} & 8.98 & 14.10 & \cellcolor{gray!20}7.30 \\
        RbA~\cite{nayal2023rba}                  & \xmark    & 73.00 & 100.0 & \cellcolor{gray!20}1.64 && 21.84 & 78.58 & 2.75 & 17.16 & \cellcolor{gray!20}2.16 \\
        UEM*~\cite{nayal2024likelihood} & \cmark   & 95.80 & 26.37 & \cellcolor{gray!20}{6.78} && 47.72 & 79.24 & 12.49 & 37.81 &  \cellcolor{gray!20}{9.90}  \\
        Ours & \cmark & \textbf{97.85} & \textbf{9.60} & \cellcolor{gray!20}\textbf{10.68} && \textbf{49.34} & 78.89 & \textbf{13.46} & \textbf{38.93} & \cellcolor{gray!20}\textbf{10.62} \\
        \bottomrule
    \end{tabular}}
    \label{tab:val-results}
\end{table*}

\begin{table*}[ht]
    \centering
    \caption{Anomaly Segmentation Performance of the Test Set of STU benchmark. Results are taken directly from~\cite{nekrasov2025stu}. All methods use the Mask4Former~\cite{yilmaz2024mask4former} architecture.}
    \scriptsize
    \resizebox{\textwidth}{!}{
    \begin{tabular}{lccccccccccc}
        \toprule
        \multirow{2}{*}{Method} & \multirow{2}{*}{\shortstack{Auxiliary \\ OOD Data}} & \multicolumn{3}{c}{Point-Level OOD} && \multicolumn{5}{c}{Object-Level OOD} \\
        \cline{3-5}
        \cline{7-11}
         &  & AUROC~$\uparrow$ & FPR@95~$\downarrow$ & AP~$\uparrow$ && RecallQ~$\uparrow$ & SQ~$\uparrow$ & RQ~$\uparrow$ & UQ~$\uparrow$ & PQ~$\uparrow$ \\
        \midrule
        Deep Ensemble~\cite{lakshminarayanan2017deepensemble}         & \xmark  & 86.74 & 58.05 & \cellcolor{gray!20}{5.17} && 16.75 & {84.49} & 10.43 & 14.16 & \cellcolor{gray!20}{8.81} \\
        MC Dropout~\cite{srivastava2014mcdropout}       & \xmark  & 61.51 & 82.37 &  \cellcolor{gray!20}{0.11} && 
        2.25 & \textbf{86.72} & 1.95 & 2.14 & \cellcolor{gray!20}{1.86} \\
        Max Logit~\cite{hendrycks2018baseline}        & \xmark  & 84.53 & 81.49 & \cellcolor{gray!20}{0.95} && 26.14 & 83.06 & 2.13 & 21.71 & \cellcolor{gray!20}{1.77} \\
        Void Classifier~\cite{blum2021fishyscapes}  & \cmark  & {85.99} & {78.60} & \cellcolor{gray!20}{{3.92}} && 17.64 & 84.40 & {8.19} & 14.89 & \cellcolor{gray!20}{{6.91}} \\
        RbA~\cite{nayal2023rba}             & \xmark  & 66.38 & 100.0 & \cellcolor{gray!20}{0.81} && {24.04} & 83.28 & 3.23 & {20.02} & \cellcolor{gray!20}{2.69} \\
        Ours   & \cmark  & \textbf{96.26} & \textbf{21.69} & \cellcolor{gray!20}{\textbf{10.17}} && \textbf{33.98} & 83.39 & \textbf{11.64} & \textbf{28.34} & \cellcolor{gray!20}{\textbf{9.71}} \\
        \bottomrule
    \end{tabular}}
    \label{tab:test-results}
\end{table*}

\section{Experiments}

\subsection{Datasets}
\paragraph{Spotting the Unexpected (STU).}
The STU benchmark~\cite{nekrasov2025stu} is a recent large-scale dataset for anomaly segmentation in 3D LiDAR. 
It contains 72 driving sequences collected with a 128-beam sensor, including both naturalistic anomalies encountered in traffic and controlled placements of diverse OOD objects such as buckets, chairs, and surfboards. 
All LiDAR points are annotated as \emph{inlier}, \emph{anomaly}, or \emph{unlabeled}. 
The benchmark provides 19 validation sequences and 51 held-out test sequences, along with 2 additional scans for closed-set training and validation. 

\paragraph{SemanticKITTI.}
We further evaluate our method on SemanticKITTI~\cite{behley2019semantickitti}, a widely used real-world LiDAR segmentation benchmark. 
We simulate OOD detection by treating outlier objects (\emph{other-structure}, \emph{other-object}) as OOD classes, while training on closed-set categories. 
Evaluation is performed on the official validation set, enabling a complementary test of OOD generalization in a controlled setting.

\subsection{Evaluation Protocol}
Following Nekrasov~\etal~\cite{nekrasov2025stu}, we report both point-level and object-level metrics.  
Point-level evaluation uses AUROC, FPR@95 and AP,  
while object-level evaluation follows the panoptic segmentation protocol, 
reporting Recall Quality (RecallQ), Segmentation Quality (SQ), Recognition Quality (RQ), Panoptic Quality (PQ), and Unknown Quality (UQ).

\subsection{Implementation Details}

We initialize the model from a Mask4Former checkpoint pretrained on SemanticKITTI~\cite{behley2019semantickitti} and Panoptic-CUDAL~\cite{tseng2025panopticcudal}, and apply the proposed Point Raise and Relative Energy Learning (REL) techniques using only a single additional sequence from the STU closed-set training set.

The model is optimized using AdamW with a learning rate of $2\times10^{-4}$ and a batch size of 8, trained for 30 epochs on NVIDIA A100 GPUs. To mitigate the imbalance caused by the limited availability of OOD objects, we set the weighting factor $\omega$ in the REL objective to 100, promoting more balanced optimization between inlier and OOD signals.
For Point Raise, we set the pull factor $\gamma$ to 2. The radius and height ranges are both set to $[0.25, 0.75]$, allowing the synthesis of diverse realistic pseudo-OOD objects.

During inference, the relative energy $\Delta E$ serves as the OOD indicator, while segmentation outputs are obtained from the Mask4Former decoder. For object-level evaluation, OOD points are clustered into instance masks using DBSCAN~\cite{dbscan}.

\begin{table}[t]
    \centering
    \footnotesize
    \caption{Anomaly Segmentation Performance on SemanticKITTI Outlier Classes. Object-level performance is not applicable because outlier classes do not have instance masks. * denotes model using Cylinder3D~\cite{zhou2020cylinder3d} architecture. Other methods use the Mask4Former~\cite{yilmaz2024mask4former} architecture.}
    \begin{tabular}{lcrrr}
        \toprule
        Method & Aux. Data & AUROC~$\uparrow$ & FPR@95~$\downarrow$ & AP~$\uparrow$ \\
        \midrule
        Max Logit~\cite{hendrycks2018baseline} & \xmark & 90.73 & 47.63 & 53.86\\
        RbA~\cite{nayal2023rba} & \xmark & 78.86 & 100.00 & 55.23 \\
        UEM~\cite{nayal2024likelihood} & \cmark & 93.15 & 37.07 & 61.73 \\
        REAL*~\cite{cen2021real} & \cmark & 84.90 & - & 20.08 \\
        APF*~\cite{apf2023} & \xmark & 85.60 & - & 36.10 \\
        LiON*~\cite{lion2025} & \cmark & 92.69 & - & 44.68 \\
        Ours & \cmark & \textbf{96.76} & \textbf{18.66} & \textbf{67.32} \\
        \bottomrule
    \end{tabular}
    \label{tab:semantic_KITTI_ood}
\end{table}

\section{Results}
\subsection{Anomaly Segmentation Performance}
\paragraph{Results on STU Benchmark.}

\cref{tab:val-results} and~\ref{tab:test-results} report anomaly segmentation results on the STU validation and test sets. The STU benchmark focuses on detecting rare and potentially hazardous objects that fall outside the training distribution. As such, the anomaly segmentation task can be naturally formulated as a \emph{point-wise OOD detection} problem, where the goal is to identify individual points that belong to unknown or unseen object categories.

Our method consistently outperforms prior approaches on both point-level detection and object-level mask quality.  
On the validation set, it attains the highest point-level AUROC (97.85), the lowest FPR@95 (9.60), and the best average precision (10.68), surpassing strong baselines reported by~\cite{nekrasov2025stu} as well as the reproduced SOTA method leveraging auxiliary OOD data~\cite{nayal2024likelihood}. Notably, Deep Ensemble~\cite{lakshminarayanan2017deepensemble}, which requires ensembling three independently trained models, achieves competitive results but at a much higher computational cost. At the object level, our method obtains the highest panoptic quality (10.62), reflecting accurate object discovery and segmentation quality. These improvements generalize well to the test set, where our approach continues to outperform all prior methods in both point-level and object-level metrics, thereby demonstrating the robustness and effectiveness of our relative energy-based learning.

\paragraph{Results on SemanticKITTI.}
\cref{tab:semantic_KITTI_ood} presents OOD detection results on the outlier classes of SemanticKITTI.
Our method achieves the best overall performance across all evaluation metrics.
Specifically, REL improves AUROC to 96.76, outperforming UEM~\cite{nayal2024likelihood} and Max Logit~\cite{hendrycks2022streethazards} by 3.6\% and 6\%, respectively.
It also reduces FPR@95 to 18.66, a substantial drop compared to UEM (37.07) and MaxLogit (47.63), indicating fewer false positives on in-distribution points.
In terms of AP, REL reaches 67.32, exceeding the strongest baseline (UEM at 61.73) by nearly 6 points.
These results demonstrate that REL provides more reliable OOD detection than both post-hoc scoring methods (Max Logit, RbA) and training-based approaches such as UEM. We are not able to report object-level performance, because the outlier classes in SemanticKITTI lack instance mask annotations.

\begin{table}[ht]
    \centering
    \caption{OOD detection performance with and without backbone finetuning. REL remains effective with a frozen backbone, and full finetuning further improves all metrics.}
    \small
    \begin{tabular}{lrrr}
        \toprule
        Training Strategy & AUROC~$\uparrow$ & FPR@95~$\downarrow$ & AP~$\uparrow$ \\
        \midrule
        OOD projector Only & 94.43 & 31.55 & 4.04 \\
        Finetune & \textbf{97.85} & \textbf{9.60} & \textbf{10.68} \\
        \bottomrule
    \end{tabular}
    \label{tab:ftandfreeze}
\end{table}

\begin{figure*}[ht]
    \centering
    \includegraphics[width=\linewidth]{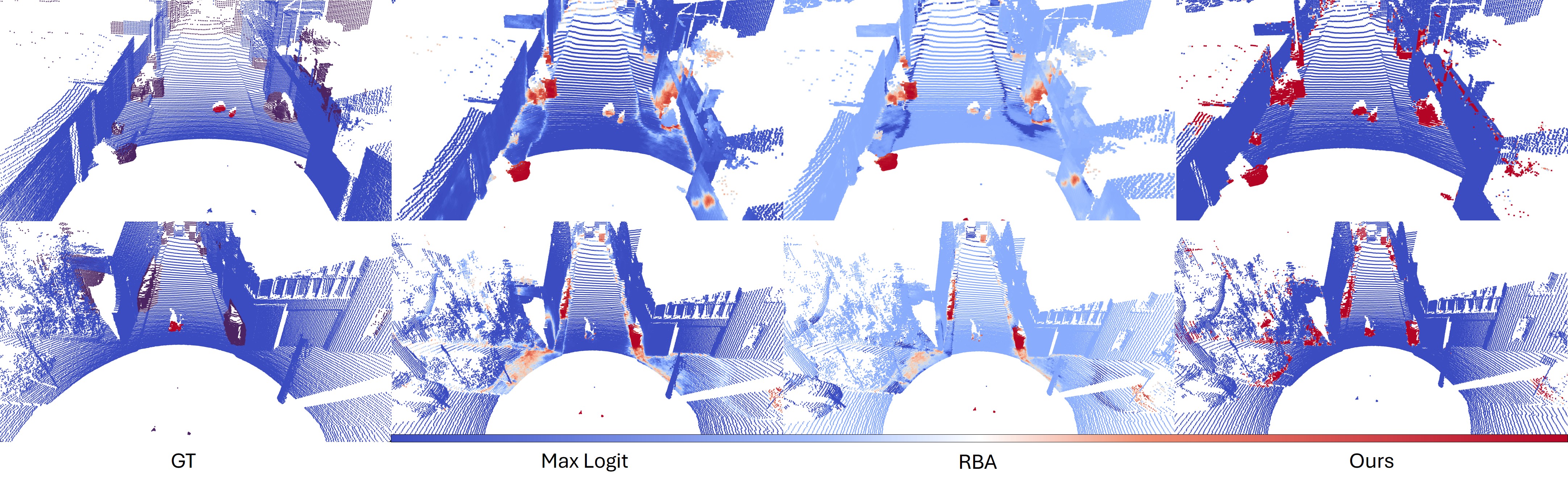}
    \caption{Visualization of anomaly segmentation results on the STU benchmark.
Points are categorized into three types: \textcolor{blue}{inlier}, \textcolor{red}{anomaly}, and \textcolor[rgb]{0.4,0,0.4}{unlabeled}. 
The unlabeled class covers a wide range of objects that may appear in driving scenes but are excluded from training and evaluation, such as rubbish bins. 
These objects are OOD relative to the closed-set training data, yet they are unlikely to pose a threat to the vehicle. 
We use a continuous gradient (color bar) to visualize each point’s predicted likelihood of belonging to \textcolor[rgb]{0,0,0.9}{ID} or \textcolor[rgb]{0.8,0,0}{OOD}.
Our method produces high-quality anomaly segmentation while keeping the false positive rate on inliers low. }
    \label{fig:vis_main}
\end{figure*}

\begin{table}[htbp]
\centering
\caption{Closed-Set segmentation performance on the validation sets of STU~\cite{nekrasov2025stu} and SemanticKITTI~\cite{behley2019semantickitti}.}
\label{tab:mean-pq}
\footnotesize
\begin{tabular}{lcc}
\toprule
Method & STU (PQ) & SemanticKITTI (PQ) \\
\midrule
Mask4Former-closed-set~\cite{marcuzzi2023maskpls} & 52.73 & 60.72 \\
Mask4Former-void~\cite{blum2021fishyscapes} & 26.96 & 47.97 \\
Mask4Former-REL & 48.98 & 59.09 \\
\bottomrule
\end{tabular}

\end{table}

\subsection{Closed-set Segmentation Performance}
\cref{tab:mean-pq} shows that Mask4Former-REL achieves comparable panoptic quality (PQ) to standard Mask4Former training on both STU and SemanticKITTI, while largely outperforming the void-classifier~\cite{blum2021fishyscapes} style OOD learning.
REL preserves strong performance across key classes without degrading segmentation quality. This confirms that REL enables reliable OOD detection with minimal impact on closed-set performance.

In addition, as shown in~\cref{tab:ftandfreeze}, REL allows training only the OOD projector while keeping the backbone frozen. This strategy preserves closed-set performance and still yields strong OOD detection, achieving AUROC of 94.43. Full finetuning further improves detection, with AUROC of 97.85 and FPR@95 reduced to 9.60.
\subsection{Qualitative Results}

\cref{fig:vis_main} presents anomaly segmentation results on the STU benchmark.
Compared to Max Logit and RbA, our method detects more complete and accurate anomaly regions while largely reducing false positives on inlier points.
Both baselines tend to produce noisy or fragmented predictions, particularly around occluded or high-curvature surfaces.
In contrast, our model yields precise and compact anomaly masks that align closely with the ground truth.
Importantly, the STU dataset includes a broad category of unlabeled points that are not supervised during training but still represent out-of-distribution content.
Our method can also successfully identify many of these unlabeled OOD points.
This shows that our method can generalize beyond labeled anomaly instances and detect a wider range of unexpected objects in open-world driving scenes.

\begin{table}[t]
    \centering
    \footnotesize
    \caption{Ablation Study on the Point Raise Hyperparameter $\gamma$. Rows marked with * indicate that Point Raise is not applied, and unlabeled points are used as auxiliary OOD data for training.}
    \begin{tabular}{lrrr}
        \toprule
        $\gamma$ & AUROC~$\uparrow$ & FPR@95~$\downarrow$ & AP~$\uparrow$ \\
        \midrule
        1 & 97.14 & 14.18 & 6.57 \\
        2 & \textbf{97.85} & \textbf{9.60} & \textbf{10.68} \\
        4 & 97.68& 9.63 & 9.42\\
        8 & 97.19& 13.88 & 8.01 \\
        None* & 90.20 & 38.31 & 1.52 \\
        \bottomrule
    \end{tabular}
    \label{tab:gamma}
\end{table}

\subsection{Ablation Study}
\paragraph{Ablation Study of the Point Raise Algorithm.}
We conduct an ablation study to investigate the impact of the pull factor $\gamma$ in the Point Raise algorithm. As shown in~\cref{tab:gamma}, the application of Point Raise consistently improves OOD segmentation performance compared to the baseline without OOD synthesis (“None”). When Point Raise is applied, a moderate value ($\gamma$ = 2) yields the best overall results, with the highest AUROC (97.85) and AP (10.68), while also achieving the lowest FPR@95 (9.60). Smaller $\gamma$ ($\gamma$ = 1) produces a bumpy surface, which still improves performance but less effectively. The higher $\gamma$ values ($\gamma$ = 4 or 8) lead to flatter perturbations, resulting in a slight decrease in performance compared to $\gamma$ = 2, although they still substantially outperform the baseline. This confirms that the Point Raise algorithm is not sensitive to the choice of hyperparameter.    

\begin{table}[t]
    \footnotesize
    \centering
    \caption{Ablation Study: Energy-based OOD Learning Objectives with our data synthesis.}
    \begin{tabular}{lrrr}
        \toprule
        Method & AUROC~$\uparrow$ & FPR@95~$\downarrow$ & AP~$\uparrow$ \\
        \midrule
        Hinge Energy~\cite{energy} & 95.77 & 20.65 & 5.31 \\
        VOS Energy~\cite{du2021vos} & 95.71 & 25.07 & 9.81 \\
        Dual Energy~\cite{dual_energy} & 96.53 & 19.45 & 5.94 \\
        Relative Energy (ours) & \textbf{97.85} & \textbf{9.60} & \textbf{10.68} \\
        \bottomrule
    \end{tabular}
    \label{tab:energy_loss}
\end{table}

\paragraph{Ablation Study of Energy-based Learning Objectives.}
We also compare different energy-based OOD learning objectives when combined with Point Raise, as shown in~\cref{tab:energy_loss}. Among existing training objectives, Dual Energy achieves relatively better AUROC (96.53) and relatively low FPR@95 (19.45), while Hinge Energy and VOS Energy loss perform worse, especially in terms of FPR@95 (20.65 and 25.07, respectively). While VOS Energy provides a good AP (9.81), its false positive rate remains higher than others. Our proposed Relative Energy objective outperforms all baselines by a clear margin, achieving the best AUROC (97.85), the lowest FPR@95 (9.60), and the highest AP (10.68). These results demonstrate that Relative Energy improves separability between in-distribution and OOD samples compared to other energy-based methods.

\subsection{Limitations}
While our method largely improves anomaly AP and reduces false positives compared to existing baselines, some challenges remain.
As shown in~\cref{fig:vis_main}, all methods, including ours, tend to struggle in regions with complex geometry, such as high-curvature surfaces or object boundaries, where shape discontinuities and sparse sampling make reliable anomaly segmentation difficult.
Although REL largely improves OOD detection performance, further work is needed to better handle ambiguous boundary cases.

\section{Conclusion}
We present \emph{Relative Energy Learning (REL)}, a simple yet effective method for point-wise OOD detection in LiDAR panoptic segmentation. REL exploits relative energy margins between positive and negative logits for end-to-end OOD training. To address the lack of OOD data, we propose \emph{Point Raise}, a geometry-aware synthesis method that generates realistic pseudo-OOD data. Our method largely improves OOD detection metrics in the STU~\cite{nekrasov2025stu} and SemanticKITTI~\cite{behley2019semantickitti}, while preserving closed-set accuracy. We expect this work to encourage further research on reliable 3D LiDAR perception in autonomous driving.

\section*{Acknowledgments}
\noindent The first two authors acknowledge the financial support from The University of Melbourne through the Melbourne Research Scholarship. This research is supported by The University of Melbourne’s Research Computing Services and the Petascale Campus Initiative.

{\small
\bibliographystyle{ieee_fullname}
\bibliography{egbib}
}

\clearpage
\setcounter{page}{1}
\maketitlesupplementary

\section{Additional Qualitative Results}
\begin{figure}[ht]
    \includegraphics[width=\linewidth]{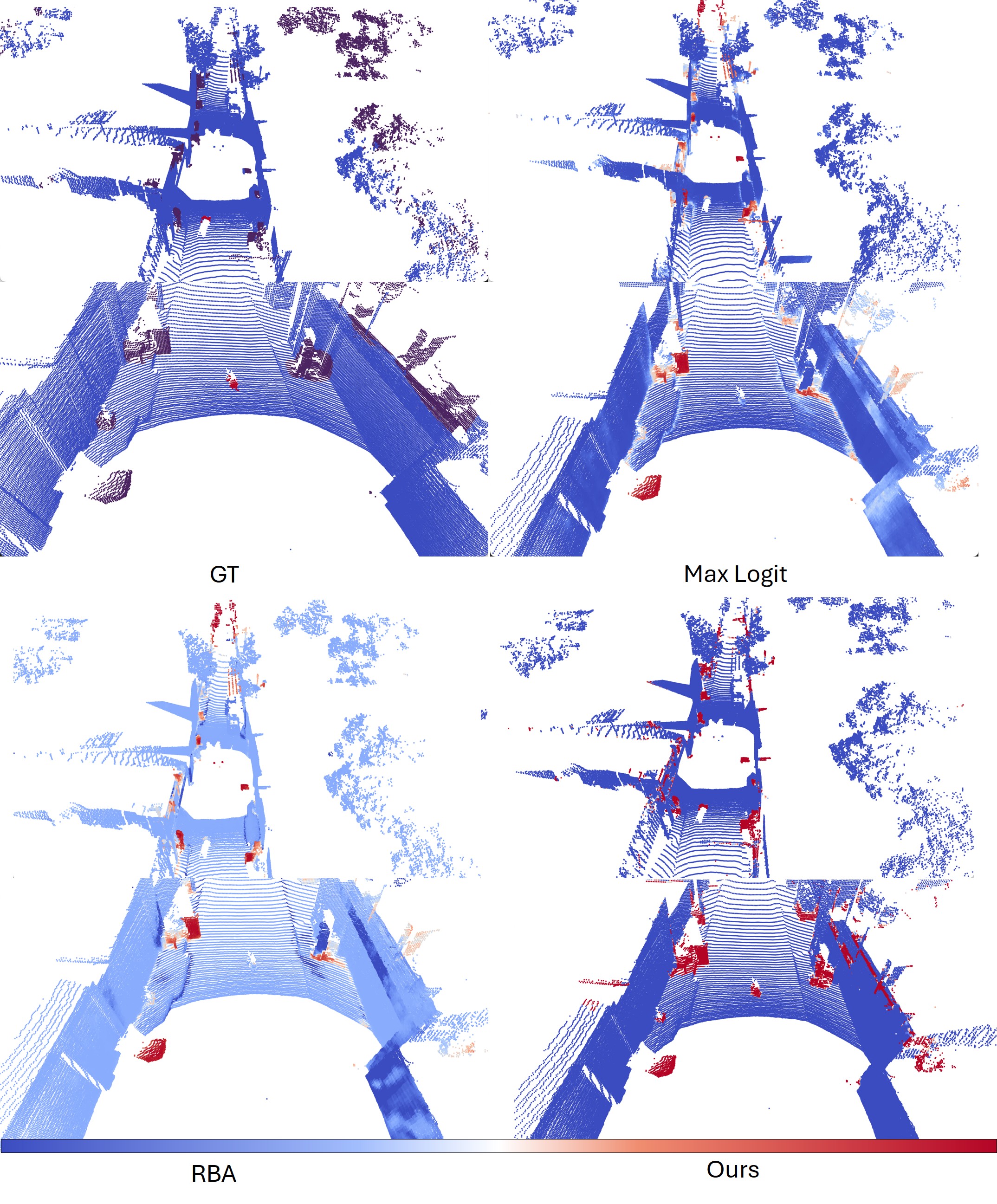}
    \caption{Visualization of anomaly segmentation results on the STU benchmark.
Points are categorized into three types: \textcolor{blue}{inlier}, \textcolor{red}{anomaly}, and \textcolor[rgb]{0.4,0,0.4}{unlabeled}. For visualization, predictions are shown with a continuous gradient from \textcolor[rgb]{0,0,0.9}{ID} to \textcolor[rgb]{0.8,0,0}{OOD}.}
    \label{fig:vis_supp}
\end{figure}

\begin{figure}[ht]
    \includegraphics[width=\linewidth]{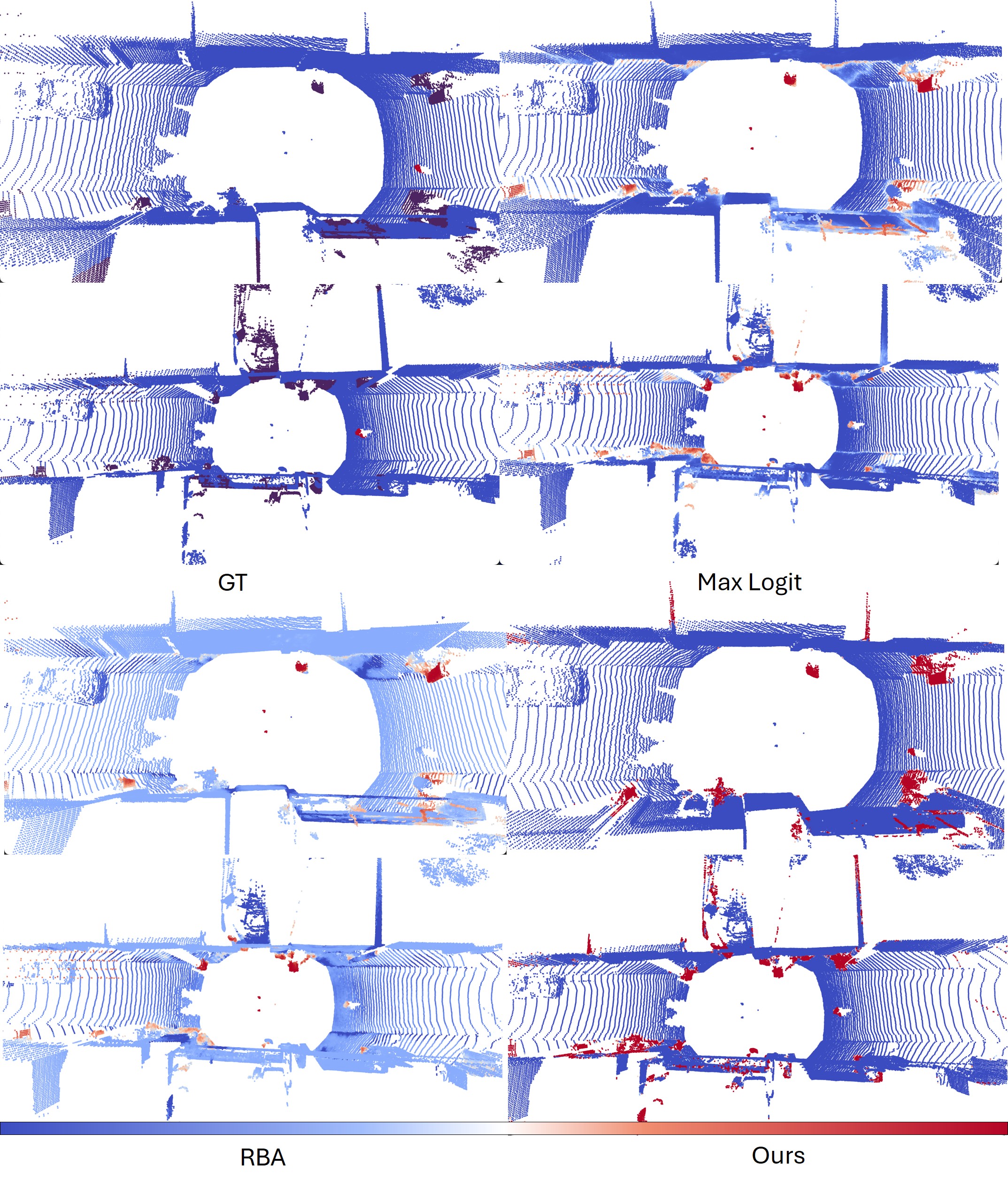}
    \caption{Visualization of anomaly segmentation results on the STU benchmark in bird's-eye view.
Points are categorized into three types: \textcolor{blue}{inlier}, \textcolor{red}{anomaly}, and \textcolor[rgb]{0.4,0,0.4}{unlabeled}. For visualization, predictions are shown with a continuous gradient from \textcolor[rgb]{0,0,0.9}{ID} to \textcolor[rgb]{0.8,0,0}{OOD}.}
    \label{fig:vis_bev}
\end{figure}

We provide additional visualization results in \Cref{fig:vis_supp} and \ref{fig:vis_bev}. In both the driver’s view and the bird’s-eye view, our proposed baseline consistently outperforms the baselines. Although our method yields more apparent false positives, these cases in fact correspond to unlabeled ground-truth points. Such detections are still valuable and essential for autonomous driving, particularly when objects are unlabeled or missing from the training dataset.

\begin{table*}[ht]
\setlength\tabcolsep{3.7pt}
    \begin{center}
    \caption{Closed-set per-class performance (Panoptic Quality) of the methods on validation sets of STU~\cite{nekrasov2025stu} dataset. With REL, our model retains comparable closed-set panoptic segmentation performance to standard Mask4Former-3D training, while largely outperforming OOD-specific methods such as the void classifier on OOD detection. The bicycle class is severely underrepresented in STU, and fine-tuning on STU leads to a noticeable drop in PQ.}
        \resizebox{\textwidth}{!}{
        \label{tab:supp-results-inlier}
        \footnotesize
        \begin{tabular}{l|c|cccccccccccccc|c}
            \toprule
            Method &
            \begin{sideways}void\end{sideways} &
            \begin{sideways}car\end{sideways} &
            \begin{sideways}truck\end{sideways} &
            \begin{sideways}bicycle\end{sideways} &
            \begin{sideways}person\end{sideways} &
            \begin{sideways}road\end{sideways} &
            \begin{sideways}sidewalk\end{sideways} &
            \begin{sideways}parking\end{sideways} &
            \begin{sideways}building\end{sideways} &
            \begin{sideways}vegetation\end{sideways} &
            \begin{sideways}trunk\end{sideways} &
            \begin{sideways}terrain\end{sideways} &
            \begin{sideways}fence\end{sideways} &
            \begin{sideways}pole\end{sideways} &
            \begin{sideways}traffic sign\end{sideways} &
            PQ \\
            \midrule
            Mask4Former~\cite{marcuzzi2023maskpls} & -- & 80.99 & 37.28 & 47.65 & 80.99 & 71.46 & 17.74 & 0.0 & 84.08 & 89.73 & 29.34 & 30.79 & 47.6 & 59.62 & 60.96 & 52.73 \\
            Mask4Former-void~\cite{blum2021fishyscapes} & 0.07 & 23.88 & 20.78 & 1.01 & 43.30 & 38.24 & 20.03 & 11.11 & 48.45 & 43.09 & 20.20 & 17.31 & 30.80 & 27.26 & 33.16 & 26.96 \\
            Mask4Former-REL & -- & 78.11 & 41.57 & 0.0 & 81.64 & 54.84 & 29.27 & 0.0 & 88.04 & 91.87 & 31.04 & 59.22 & 41.88 & 27.29 & 60.99 & 48.98 \\
            \bottomrule
        \end{tabular}
        }
    \end{center}
    \vspace{-0.5cm}
\end{table*}

\begin{table*}[ht]
\setlength\tabcolsep{3.7pt}
    \begin{center}
    \caption{Closed-set per-class performance (Panoptic Quality) of the methods on validation sets of SemanticKITTI~\cite{behley2019semantickitti}.}
        \resizebox{\textwidth}{!}{
        \label{tab:supp-results-semKITTI}
        \footnotesize
        \begin{tabular}{l|c|ccccccccccccccccccc|c}
            \toprule
            Method &
            \begin{sideways}void\end{sideways} &
            \begin{sideways}car\end{sideways} &
            \begin{sideways}truck\end{sideways} &
            \begin{sideways}bicycle\end{sideways} &
            \begin{sideways}motorcycle\end{sideways} &
            \begin{sideways}other vehicle\end{sideways} &
            \begin{sideways}person\end{sideways} &
            \begin{sideways}bicyclist\end{sideways} &
            \begin{sideways}motorcyclist\end{sideways} &
            \begin{sideways}road\end{sideways} &
            \begin{sideways}sidewalk\end{sideways} &
            \begin{sideways}parking\end{sideways} &
            \begin{sideways}other ground\end{sideways} &
            \begin{sideways}building\end{sideways} &
            \begin{sideways}vegetation\end{sideways} &
            \begin{sideways}trunk\end{sideways} &
            \begin{sideways}terrain\end{sideways} &
            \begin{sideways}fence\end{sideways} &
            \begin{sideways}pole\end{sideways} &
            \begin{sideways}traffic sign\end{sideways} &
            PQ \\
            \midrule
            Mask4Former~\cite{marcuzzi2023maskpls} & -- & 93.53 & 59.39 & 62.55 & 64.82 & 54.36 & 79.61 & 89.16 & 25.01 & 93.24 & 77.90 & 28.79 & 0.0 & 87.27 & 87.28 & 51.08 & 59.92 & 24.85 & 56.76 & 58.14 & 60.72\\
            Mask4Former-void~\cite{blum2021fishyscapes} & 6.08 & 74.36 & 47.00 & 32.19 & 43.34 & 33.30 & 42.90 & 68.75 & 00.33 & 93.35 & 77.07 & 19.01 & 0.0 & 82.77 & 81.34 & 47.56 & 56.94 & 19.98 & 54.48 & 36.82 & 47.97 \\
            Mask4Former-REL & -- & 93.36 & 73.04 & 58.37 & 64.87 & 56.61 & 77.91 & 86.64 & 0.0 & 92.27 & 76.27 & 21.41 & 0.0 & 85.65 & 88.01 & 52.07 & 60.37 & 21.30 & 56.78 & 57.77 & 59.09 \\
            \bottomrule
        \end{tabular}
        }
    \end{center}
    \vspace{-0.75cm}
\end{table*}

\end{document}